# Research and Applications

# Towards objective and systematic evaluation of bias in artificial intelligence for medical imaging

Emma A.M. Stanley , BASc[1,2,3,4,*], Raissa Souza , BSc[1,2,3,4], Anthony J. Winder, MSc[2,3],
Vedant Gulve[2], Kimberly Amador, BSc[1,2,3,4], Matthias Wilms, PhD[3,4,5,6],
Nils D. Forkert , PhD[2,3,4,6,7,8]

[1]Biomedical Engineering Graduate Program, University of Calgary, Calgary, Alberta, T2N 1N4, Canada, [2]Department of Radiology, University of Calgary, Calgary, Alberta, T2N 4N1, Canada, [3]Hotchkiss Brain Institute, University of Calgary, Calgary, Alberta, T2N 4N1, Canada, [4]Alberta Children's Hospital Research Institute, University of Calgary, Calgary, Alberta, T2N 4N1, Canada, [5]Department of Pediatrics, University of Calgary, Calgary, Alberta, T2N 4N1, Canada, [6]Department of Community Health Sciences, University of Calgary, Calgary, Alberta, T2N 4N1, Canada, [7]Department of Clinical Neuroscience, University of Calgary, Calgary, Alberta, T2N 4N1, Canada, [8]Department of Electrical and Software Engineering, University of Calgary, Calgary, Alberta, T2N 1N4, Canada

*Corresponding author: Emma A.M. Stanley, BASc, Biomedical Engineering Graduate Program, University of Calgary, 3330 Hospital Drive NW, Calgary, Alberta, T2N 4N1, Canada (emma.stanley@ucalgary.ca)

M. Wilms and N.D. Forkert shared senior authorship.

## Abstract

**Objective:** Artificial intelligence (AI) models trained using medical images for clinical tasks often exhibit bias in the form of subgroup performance disparities. However, since not all sources of bias in real-world medical imaging data are easily identifiable, it is challenging to comprehensively assess their impacts. In this article, we introduce an analysis framework for systematically and objectively investigating the impact of biases in medical images on AI models.

**Materials and Methods:** Our framework utilizes synthetic neuroimages with known disease effects and sources of bias. We evaluated the impact of bias effects and the efficacy of 3 bias mitigation strategies in counterfactual data scenarios on a convolutional neural network (CNN) classifier.

**Results:** The analysis revealed that training a CNN model on the datasets containing bias effects resulted in expected subgroup performance disparities. Moreover, reweighing was the most successful bias mitigation strategy for this setup. Finally, we demonstrated that explainable AI methods can aid in investigating the manifestation of bias in the model using this framework.

**Discussion:** The value of this framework is showcased in our findings on the impact of bias scenarios and efficacy of bias mitigation in a deep learning model pipeline. This systematic analysis can be easily expanded to conduct further controlled in silico trials in other investigations of bias in medical imaging AI.

**Conclusion:** Our novel methodology for objectively studying bias in medical imaging AI can help support the development of clinical decision-support tools that are robust and responsible.

## Background and significance

Artificial intelligence (AI) systems are often praised for having the potential to serve as an objective decision-maker in health care. However, recent studies have shown that AI can produce systematic disparities in performance outcomes between subgroups; such models are considered to be biased or unfair.[1–5] This is of particular concern in the domain of AI for medical imaging, where the first systems have been approved for assisting clinicians in making diagnostic and treatment decisions for their patients.[6] In this context, models may overfit to majority populations in the training data or learn spurious correlations that lead to poor generalization ability. For instance, various studies have found that deep learning models trained on clinical tasks achieve better performance on racial subgroups that represent the majority in the training data.[2,3,7] Other studies have shown that models trained on medical images can encode explicit information about image acquisition systems or sociodemographic characteristics.[8–10] This suggests that models could learn these "bias attributes" rather than representations of the disease of interest itself, which may lead to so-called shortcut learning.[11] Given these concerning findings, a considerable amount of recent research has focused on applying and developing strategies to mitigate subgroup performance disparities in clinical models. Noteworthy examples include mitigating racial bias in cardiac segmentation,[2] removing confounders in HIV diagnosis from brain scans,[12] pruning to debias chest X-ray[13] and dermatological models,[14] and harmonizing information related to magnetic resonance (MR) scanners encoded in models.[15]

Moreover, recent studies have started investigating how biases from medical imaging data are encoded in models.[16–18]

An improved understanding of this holds great value since it could facilitate the development of models that are inherently robust to such biases. However, it is difficult to comprehensively and objectively assess this phenomenon because medical imaging datasets contain numerous biases or spurious correlations that deep learning models may use, but that we as humans do not know about, cannot fully understand, or have no ability to disentangle.[11] The same is true for studies that propose or benchmark bias mitigation methods[19,20]—while it is valuable to know that certain techniques are successful in reducing performance disparities between subgroups on certain clinical tasks, it remains unknown if such strategies are correcting for the true sources of bias in images, and if they are also effective when applied to data from populations not represented in the datasets used to evaluate these mitigation strategies. Therefore, to objectively and systematically study how biases manifest in deep learning models and confidently evaluate the robustness of bias mitigation strategies, it is imperative to understand and control the biases in a medical imaging dataset in the first place.

In other domains of computer vision-based AI research, "toy" datasets, such as MNIST[21] and its variations,[22,23] enable researchers to benchmark deep learning architectures and study simple, controlled scenarios to evaluate how trained models perform on data with, for example, intentional biases such as background color. However, there is a substantial gap between understanding how bias in these rather simple and small (28 × 28 pixels) 2-dimensional images impact deep learning models, compared to complex interactions between anatomical and scanner-induced biases in medical images that can comprise millions of voxels. In this work, we specifically aim to bridge this gap by introducing a method that integrates the customization and control of MNIST-like datasets with the scale and complexity of 3-dimensional medical images.

## Objective

The purpose of this study is to introduce and utilize an analysis framework that enables machine learning researchers to conduct systematic investigations of how biases that are relevant in medical imaging (eg, morphology changes) impact deep learning pipelines. As a key component, we use the recently proposed Simulated Bias in Artificial Medical Images (SimBA)[24] tool for generating structural neuroimaging datasets with known, user-defined biases. In this framework, these biases are not intended to faithfully emulate any specific real-world sociodemographic subpopulation, since the imaging features that introduce bias within these subpopulations are often complex, interacting, and/or unknown. Instead, our simulated biases introduce hypothetical subgroups in a dataset that present with defined confounding features, which may result in shortcut learning within medical imaging AI models.

Our specific contributions include demonstrating the controlled and customizable nature of this data-generating mechanism to generate counterfactual subject-paired synthetic datasets with simulated biases in different spatial locations of the human brain, or without the presence of bias at all. Then, with identical model training schemata, we conduct in silico trials exploring bias in AI, in which we systematically study the impact of different bias manifestations on the same deep learning pipeline. Moreover, we use these synthetic datasets to assess how well existing bias mitigation techniques are able to compensate for the (simulated) biases by reducing subgroup performance disparities. These experiments exemplify how this framework can be used within the medical imaging AI research community to further investigate how bias is learned, encoded, and potentially can be mitigated in deep learning pipelines.

## Methods
### Generation and study of dataset bias

As the basis of this work, we use and extend our recently proposed SimBA[24] tool to generate synthetic T1-weighted brain MR images. Briefly described, morphological variations representing disease, bias, and subject effects derived from real-world MR datasets are introduced by deforming a template image that represents an average brain morphology. These effects/deformations were sampled from principal component analysis (PCA)-based generative models of non-linear deformations computed by non-linear registration of real-world MR datasets to the template image, such that each effect represents a particular degree of morphological variation within the expected range of inter-subject human brain anatomy. As described below, the systematic integration of these initially arbitrary deformations into the images enables the modeling of distinct simulated subjects within a dataset.

In our setup, disease and bias effects are represented by spatially localized deformations, whereas individual subjects are represented by variation in global brain morphology. The procedure of applying morphological effects to the template image is repeated numerous times to generate a large dataset, where the user specifies the images belonging to each target (disease) class and the images containing bias effects. Thus, each image in the resulting dataset contains a unique combination of a distinct global subject morphology, localized effects that determine "disease" or "non-disease" status, and the potential presence of additional localized effects that determine bias subgroup membership. The whole process is user-defined such that the dataset contains a specified number of images, number of target classes, and proportions of images in each class with or without bias effects, which enables a controlled setup for investigating the impact of a known bias on deep learning models.

In this work, we extend this basic methodology to predefine stratified distributions of subject and disease effect magnitude. Simply put, this ensures that the subject and disease effect magnitude distributions for each class and bias group are similar, which helps to prevent these distributions from introducing an additional, unwanted source of bias in deep learning models. This significant extension of SimBA is, therefore, crucial for the aim of this work. Further details are depicted in Figure 1.

The datasets generated for this study used the SRI24 atlas[25] with LPBA40 labels[26] as the brain template, with dimensions of 173 × 211 × 155 voxels and an isotropic resolution of 1 mm³. More precisely, the morphological effects added to this template were derived from PCA models fit to stationary velocity fields of T1-weighted MRIs (Log-Euclidean framework[27]) which were obtained via non-linear diffeomorphic registration of 50 real images from the IXI[28] database to the SRI24 atlas, as described in Stanley et al.[24]

Each full dataset generated for the experiments in this work is comprised of 2002 3-dimensional images of distinct simulated subjects, with 2 target classes ("disease," $N = 1000$, and "non-disease," $N = 1002$), and 2 bias groups ("bias," containing an additional morphological effect, and "non-bias," which does not contain such an effect). 70% of the images in the disease class were defined to belong to the bias group, while 30% of the images in the non-disease class were defined to belong to the bias group. With this unequal representation of the bias group in each class, it is expected that the deep learning model uses the presence of the bias effect as a shortcut for predicting the disease class, resulting in subgroup performance disparities.

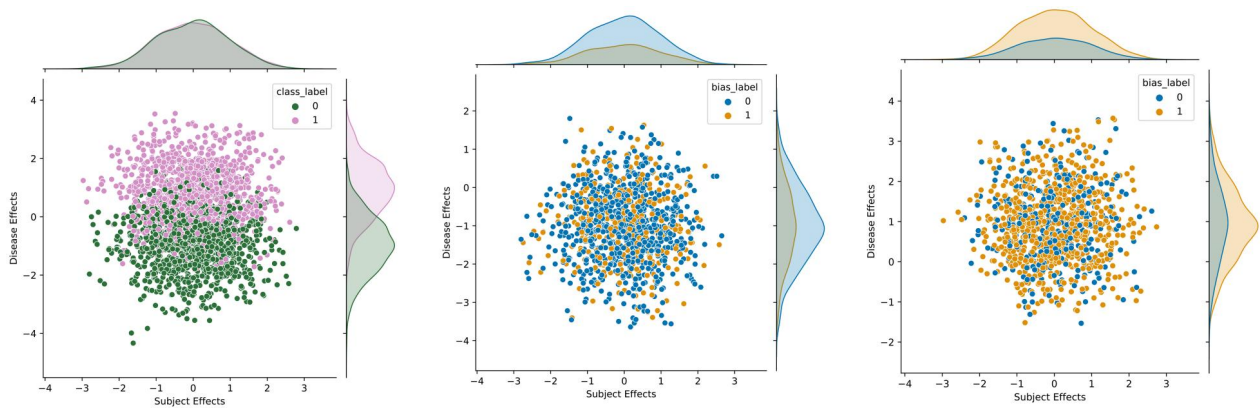

**Figure 1.** Kernel density estimate/scatter plots of subject and disease effect magnitude sampled from principal component analysis models. The distribution of subject effect magnitude within each target class, or the joint distribution of subject and disease effect magnitudes within each bias group can be identified by deep learning models and lead to performance disparities if there are differences between subgroups. Therefore, it is important to stratify such effects as shown in the figure to minimize this distribution-induced bias. The sampling distributions shown in the figure are identical for each bias scenario. Left: All samples, middle: non-disease class, right: disease class.

Table S1 provides the detailed dataset generation parameters. We defined the disease region, which contains the effects that are meant to act as the prediction target for the deep learning model, as the left insular cortex. Next, we generated datasets for 3 bias scenarios: (1) No Bias Effect, (2) "Near" Bias Effect (left putamen, adjacent to disease region), and (3) "Far" Bias Effect (right postcentral gyrus, opposite hemisphere as the disease region and on a different axial slice). These 3 bias scenarios are "paired" such that the images in each dataset contain the same subject effects and disease effects—in other words, each dataset acts as a counterfactual, where the same generated subjects have either no bias effect, or a bias effect in the near or far regions, respectively. Thus, evaluation of a model with identical weight initialization and deterministic training allows for direct comparison of the impact of each bias scenario on the deep learning pipeline. We applied this training schema for prediction of the target class from the imaging data using a convolutional neural network (CNN).

## CNN model and training

The CNN used for all experiments in this work contains 5 convolutional blocks (filters = 32, 64, 128, 256, 512) comprised of 3×3×3 convolution filters, batch normalization, sigmoid activation, and 2×2×2 max pooling, followed by average pooling, 20% dropout, a flatten layer, and a dense classification layer. A batch size of 4 and learning rate of 1e-4 was used, and models were trained until convergence, by validation loss early stopping patience = 15. For each bias scenario (No Bias, Near Bias, Far Bias), we trained the same model architecture using the same 5 weight initialization seeds with deterministic GPU state, in which each seed replicate uses the same dataset train, validation, and test splits of 50%/25%/25%. Each split was stratified by subject effects, disease effects, bias group label, and disease class label. Thus, evaluation of each bias scenario is a counterfactual of the same model and same dataset, where the difference is in the absence of the bias effect, or the bias effect in different locations. The models were implemented in Keras/Tensorflow v. 2.10 using a NVIDIA GeForce RTX 3090 GPU.

## Bias mitigation

Reweighing was implemented as proposed by Calders et al.[29] Equations used to calculate sample weight based on bias group and disease class are described in the Supplementary Material.

We implemented bias unlearning as described by Dinsdale et al,[15] where the goal is to is to modify the weights of the feature encoder (ie, the entire CNN except for the dense classification layer) such that the model is no longer able to predict the bias group while retaining the ability to predict the disease status. The same feature encoder backbone was used with 2 classification heads (ie, the final dense layer), one for the disease classification and one for the bias group classification. First, the encoder and disease prediction head were trained until convergence. Then, the encoder was frozen and the disease prediction head was replaced with a bias prediction head, which was trained to convergence. Subsequently, the unlearning process takes place, in which the cross-entropy loss for predicting the disease is minimized, while a confusion loss for predicting the bias group is maximized. This iterative training takes place until the disease prediction accuracy has stopped improving and the bias prediction accuracy has stopped decreasing. In our experiments, only 5 epochs were necessary to unlearn the bias group. Bias prediction results are reported in Table S5.

For the bias group model approach, we follow Puyol-Antón et al.[2] The model was pre-trained for 5 epochs on the full dataset. Then, using the pre-trained model as a basis, separate models were trained on both the bias and non-bias groups. Accuracy, true positive rates, and false positive rates were computed for each group model individually, and then compared to get the reported subgroup performance disparities.

## Evaluation

Model accuracy was used as a metric to determine efficacy in achieving the target prediction task of disease classification, where the disease class is a positive prediction, and the non-disease class is a negative prediction. Subgroup performance disparities, or the difference in true positive rate ($\Delta$TPR) and false positive rate ($\Delta$FPR) between the bias group and the non-bias group, were the primary outcome metric for measuring bias (Equations 1 and 2):

$$\Delta \text{TPR} = \text{TPR}_{\text{biasgroup}} - \text{TPR}_{\text{non-biasgroup}}. \tag{1}$$

$$\Delta \text{FPR} = \text{FPR}_{\text{biasgroup}} - \text{FPR}_{\text{non-biasgroup}}. \tag{2}$$

Standard deviations across 5 model seeds are reported as error values.

Exploiting the benefits of the counterfactual setup, we measured performance disparities relative to the baseline (No Bias) scenario (Equations 3 and 4), since there are minimal but non-zero performance disparities due to differences in the underlying disease and subject effect distributions. Thus, any measured relative performance disparity between bias groups can be attributed purely to the presence of the morphological bias effects.

$$\text{Relative}\Delta \text{TPR} = \Delta \text{TPR} - \Delta \text{TPR}_{\text{NoBias}}. \tag{3}$$

$$\text{Relative}\Delta \text{FPR} = \Delta \text{FPR} - \Delta \text{FPR}_{\text{NoBias}}. \tag{4}$$

## Statistics

We performed a 2-way repeated measures ANOVA on $\Delta$TPR and $\Delta$FPR with bias scenario (No Bias, Near Bias, Far Bias) and bias mitigation method (naïve, reweighing, unlearning, group models) as the factors. Following significant effects, we performed Shapiro-Wilk tests to check for normality and 2-tailed comparison tests (paired *t*-tests if normality test was passed; otherwise Wilcoxon matched-pairs signed rank test) with Bonferroni correction to compare between specific bias scenarios and mitigation methods (Tables S3 and S4). The corrected significance level was $\alpha = 0.005$. Analysis was performed in GraphPad Prism 10.0.2.

## Explainability

SmoothGrad[30] saliency maps were used to explore the extent to which the bias regions influenced the model in predicting the target class. Final saliency maps were generated from an average of 10 images from each bias group that were correctly classified as belonging to either the disease or non-disease class. Quantitative explainability is reported using weighted saliency scores,[31] which represents the relative intensity of salient voxels in a given brain region across the whole saliency map.

## Results

### Simulated bias effects lead to subgroup performance disparities in trained models

All models trained using data from the 3 bias scenarios demonstrate high overall classification accuracy, with $84.81 \pm 0.62\%$,

$87.68 \pm 0.20\%$, and $86.77 \pm 0.71\%$ for the No Bias, Near Bias, and Far Bias datasets, respectively. The presence of the bias effect resulted in significant performance disparities compared to the No Bias scenario, with a relative $\Delta$TPR of $22.12 \pm 2.72\%$ ($P < .0001$) and $\Delta$FPR of $20.32 \pm 2.81\%$ ($P < .0001$) in the Near Bias scenario, and a relative $\Delta$TPR of $17.35 \pm 2.78\%$ ($P = .0002$) and relative $\Delta$FPR of $11.88 \pm 6.22\%$ ($P = .0008$) in the Far Bias scenario. These results indicate that the models likely used the bias effect as a shortcut for predicting the (positive) disease class, since the disease class contains a higher proportion of the bias group compared to the non-disease class. However, the models may have used this shortcut learning to a lesser degree when the bias region is farther from the disease region, as evidenced by the lower relative $\Delta$FPR and $\Delta$TPR in the Far Bias scenario compared to the Near Bias scenario.

### Bias mitigation alleviates performance disparities to different extents

Each of the bias mitigated models demonstrated high overall classification accuracy, comparable to that of the naïve model (Table 1). It was also observed that all bias mitigation strategies reduce TPR and FPR disparities between the bias groups, albeit to different degrees (Figure 2).

The reweighing strategy almost perfectly mitigated any measured performance disparity between the bias groups, to respective values of relative $\Delta$TPR and relative $\Delta$FPR of $0.44 \pm 1.80\%$ and $-0.69 \pm 1.45\%$ for the Near Bias scenario, and $-0.80 \pm 0.64\%$ and $-0.32 \pm 2.13\%$ for the Far Bias scenario.

The unlearning strategy also reduced the performance disparities between groups, although to a lesser degree. More precisely, the relative $\Delta$TPR was reduced to $12.22 \pm 3.04\%$ for the Near Bias scenario and $4.60 \pm 4.20\%$ for the Far Bias scenario. The relative $\Delta$FPR was reduced to $15.51 \pm 4.87\%$ and $3.58 \pm 3.01\%$ for the Near and Far Bias scenarios, respectively. Notably, unlearning reduced performance disparities to a greater degree for the Far Bias scenario compared to the Near Bias scenario. For all bias scenarios, the ability of the model to predict the bias label was reduced to chance level (Table S5). Performing unlearning with the dataset for the No Bias scenario resulted in lower and more variable model performance across all metrics (Table 1).

Training separate models for each bias group also reduced the average $\Delta$TPR and $\Delta$FPR to near-zero, relative to the group models trained on the No Bias dataset. The relative $\Delta$TPR and $\Delta$FPR were $-2.25 \pm 6.50\%$ and $-1.47 \pm 6.45\%$ for the Near Bias scenario, and $-3.13 \pm 6.62\%$ and $-1.54 \pm 1.59\%$ for the Far Bias scenario. However, it was observed that the group model for the No Bias baseline demonstrated significant subgroup performance disparities compared to the No Bias naïve model ($\Delta$TPR = $22.86 \pm 5.52\%$, $P = .0007$ and $\Delta$FPR = $18.70 \pm 2.42\%$, $P = .0001$). Although the No Bias dataset did not have bias effects added, each bias group model was trained on a different number of images from each disease class (Table S2), resulting in a typical class imbalance problem. Thus, while the addition of the near and far bias effects did not lead to any further performance disparities into the group models, we instead observed disparities implicitly introduced due to the different levels of bias group representation in each target class (Figure S1).

Table 1. Performance of each bias mitigation strategy for each bias scenario.

| | Aggregate | | | | Bias group | | | | Non-bias group | | | |
|---|---|---|---|---|---|---|---|---|---|---|---|---|
| | Naïve | RW | UL | GM | Naïve | RW | UL | GM | Naïve | RW | UL | GM |
| **Accuracy (%)** | | | | | | | | | | | | |
| No bias | 84.81 ± 0.62 | 85.01 ± 0.39 | 79.28 ± 2.76 | | 85.24 ± 0.61 | 85.08 ± 0.64 | 79.84 ± 6.66 | 88.23 ± 0.34 | 84.38 ± 1.18 | 84.94 ± 0.44 | 78.73 ± 10.47 | 87.01 ± 0.78 |
| Near bias | 87.68 ± 0.20 | 84.33 ± 0.39 | 86.65 ± 0.72 | | 88.21 ± 0.39 | 84.68 ± 0.29 | 87.34 ± 1.05 | 88.06 ± 0.88 | 87.15 ± 0.75 | 83.98 ± 0.77 | 85.98 ± 1.07 | 87.41 ± 0.83 |
| Far bias | 86.77 ± 0.71 | 84.89 ± 0.52 | 85.29 ± 0.99 | | 87.10 ± 0.99 | 85.32 ± 0.36 | 85.48 ± 2.61 | 87.90 ± 0.40 | 86.45 ± 0.69 | 84.46 ± 0.75 | 85.10 ± 2.59 | 87.17 ± 0.71 |
| **TPR (%)** | | | | | | | | | | | | |
| No bias | 85.73 ± 1.01 | 85.00 ± 1.67 | 80.24 ± 17.02 | | 85.44 ± 0.91 | 84.78 ± 1.65 | 79.71 ± 20.57 | 93.44 ± 1.54 | 86.47 ± 1.61 | 85.59 ± 1.92 | 81.18 ± 17.96 | 70.59 ± 6.15 |
| Near bias | 88.31 ± 0.74 | 84.44 ± 1.44 | 89.60 ± 4.34 | | 93.89 ± 0.79 | 84.33 ± 1.44 | 92.67 ± 3.69 | 92.67 ± 2.56 | 73.53 ± 2.08 | 84.71 ± 1.68 | 81.47 ± 6.29 | 72.06 ± 1.80 |
| Far bias | 85.08 ± 2.53 | 85.89 ± 0.29 | 85.24 ± 6.48 | | 89.56 ± 3.10 | 85.44 ± 0.25 | 86.22 ± 5.91 | 92.67 ± 0.72 | 73.24 ± 1.92 | 87.06 ± 0.66 | 82.65 ± 8.54 | 72.94 ± 3.83 |
| **FPR (%)** | | | | | | | | | | | | |
| No bias | 16.10 ± 1.28 | 14.98 ± 1.40 | 21.67 ± 20.70 | | 15.28 ± 0.81 | 14.12 ± 3.05 | 20.29 ± 20.57 | 25.59 ± 2.87 | 16.39 ± 1.64 | 15.30 ± 0.86 | 22.19 ± 20.76 | 6.89 ± 1.26 |
| Near bias | 12.95 ± 0.95 | 15.78 ± 1.53 | 16.25 ± 3.73 | | 26.84 ± 2.21 | 14.41 ± 3.35 | 26.76 ± 6.10 | 24.12 ± 5.46 | 7.79 ± 1.63 | 16.28 ± 1.41 | 12.33 ± 3.01 | 6.89 ± 1.13 |
| Far bias | 11.55 ± 1.67 | 16.10 ± 1.21 | 14.66 ± 6.13 | | 19.41 ± 4.81 | 15.00 ± 1.61 | 16.47 ± 6.44 | 24.71 ± 1.23 | 8.63 ± 1.18 | 16.50 ± 1.18 | 13.99 ± 6.14 | 7.54 ± 1.25 |

Errors are reported as standard deviation across five model seeds.
Abbreviations: Naïve = no bias mitigation; RW = reweighing; UL = unlearning; GM = group models; TPR = true positive rate; FPR = false positive rate.

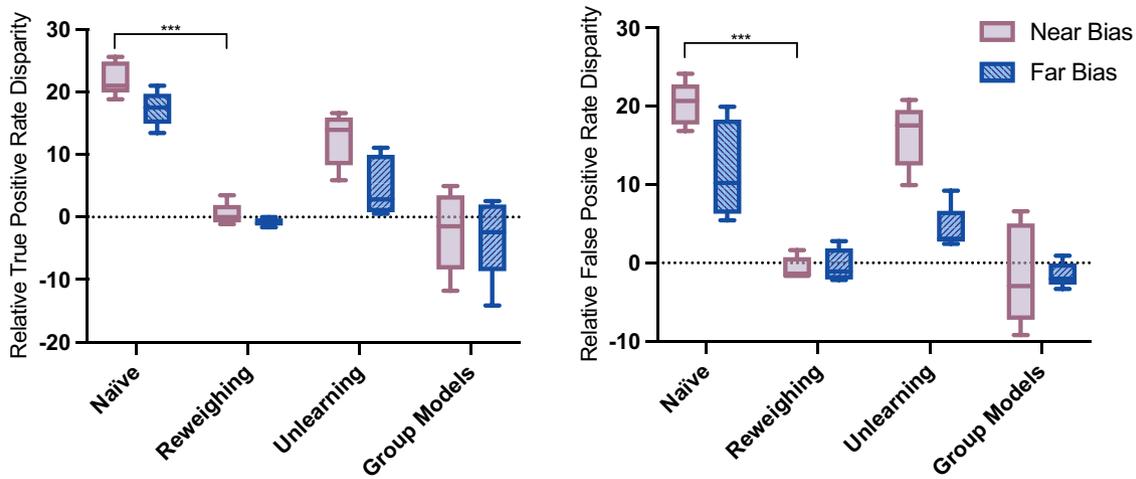

**Figure 2.** Relative true (left) and false (right) positive rate disparities between bias groups relative to the No Bias baseline for each bias mitigation strategy and bias scenario across 5 model seeds. Markers indicate significant differences between the naïve model performance (ΔTPR, ΔFPR) and performance after bias mitigation (***$P<.001$).

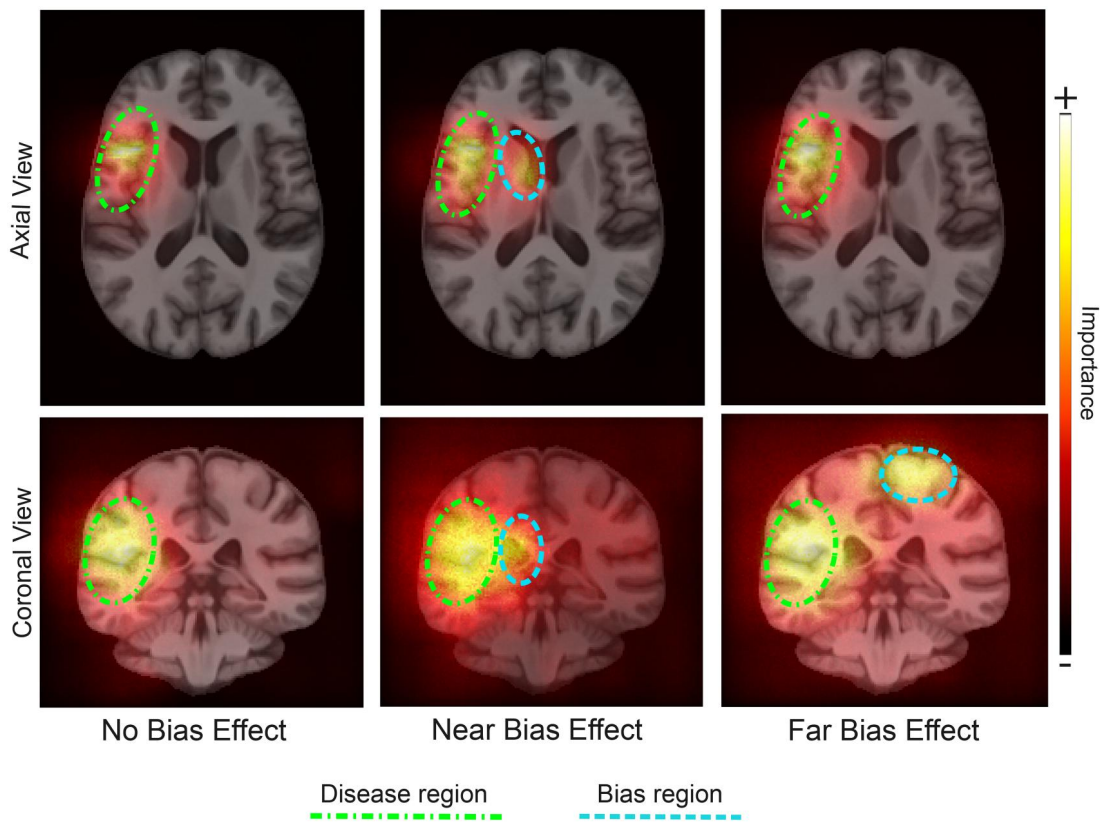

**Figure 3.** Average SmoothGrad saliency maps of correctly classified "subjects" for the bias group in the disease class, for each bias scenario in the naïve models.

## Bias effects are identified with model explainability

In the naïve model, saliency maps strongly highlighted both the disease and the corresponding bias region in each of the bias scenarios (Figure 3). Quantitatively, weighted saliency scores demonstrated a higher relative intensity of salient voxels within the regions affected by the bias effect for both the Near and Far Bias scenarios relative to the No Bias baseline (Figure 4). Furthermore, weighted saliency score results were in line with the efficacy of each bias mitigation strategy. For reweighing and group models, performance disparities were reduced to near-zero, and the corresponding saliency scores were similar to the No Bias baseline. The higher magnitude of performance disparities that remained after unlearning was reflected in the higher bias region saliency scores for this strategy.

## Discussion

Overall, the results of this study highlight the utility and benefits of using the proposed analysis framework for performing a systematic and controlled evaluation of bias in medical imaging AI. We generated counterfactual neuroimaging datasets for 3 bias scenarios using an extended version of the SimBA tool and simulated in silico trials of each dataset with identical and deterministic deep learning pipelines. Since the exact bias effects that we aim to study are known in this setup and it is known how models perform on the counterfactual dataset without bias, the naïve models provide a strong baseline for objectively interpreting results of subsequent analyses. Thus, in the evaluation of each bias mitigation strategy, we can better understand the situations these mitigation strategies are effective in, or the reasons they may fail.

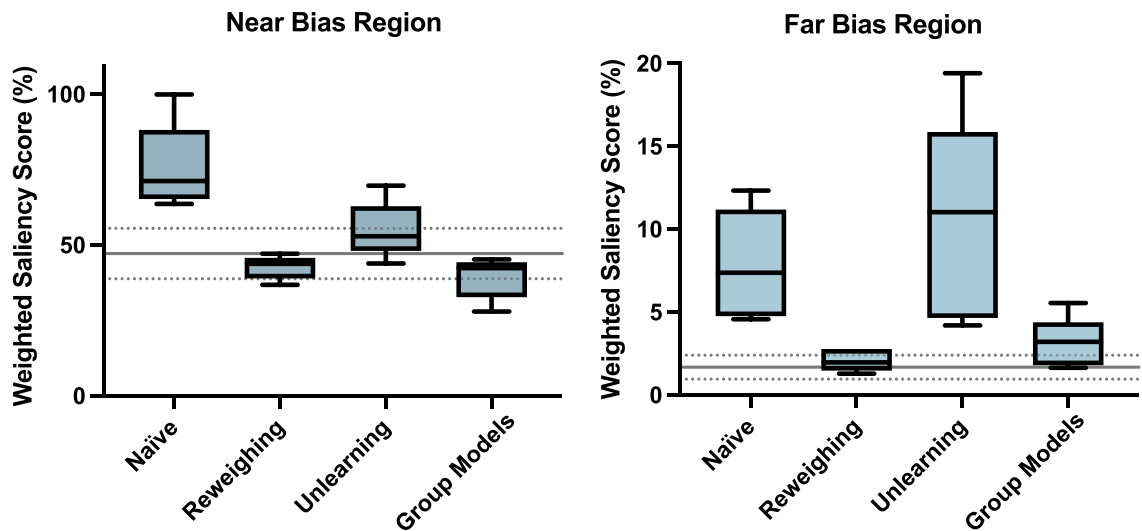

**Figure 4.** Weighted saliency scores calculated from average saliency maps over 5 model seeds for the bias group in the disease class for the brain regions most affected by "near" bias effect (left) and "far" bias effect (right). The horizontal line represents the weighted saliency score average ± standard deviation in the corresponding region for the No Bias naïve model.

The results showed that reweighing completely mitigated performance disparities caused by the single source of bias we introduced. Thus, several real-world evaluations of reweighing[2,3,32] may have been successful as a result of being applied to subgroups associated with the predominant sources of bias in the data. On the other hand, systematically introducing more complexity (eg, additional sources of bias) to these synthetic datasets may help to improve our understanding why reweighing fails to alleviate performance disparities in various other real-world datasets.[19,32–34] For the unlearning strategy, it was found that the model's ability to predict the bias label was reduced to chance level, but this did not effectively mitigate subgroup performance disparities. This may be due to the bias features and the features necessary for predicting the target class being too similar to be fully disentangled. Understanding why this occurs likely requires deeper investigation into the model's internal representations of disease and bias effects, which would be a straightforward task with these synthetic datasets since there are only 2 user-defined effects that the model can use for the prediction task. Furthermore, given the simulated counterfactual No Bias scenario, it can be seen that unlearning applied to this dataset even harms the model, resulting in lower performance on average and higher variability across model seeds. This finding is especially important considering that potential bias attributes in real-world datasets, such as scanner models or self-reported demographics, could be inaccurately labeled. Further work could be performed using this framework to analyze how mislabeling of bias attributes affects the result of this and other mitigation strategies. Moreover, the No Bias counterfactual datasets made it clear that performance disparities in the group models were not caused by the bias effects, but rather by the target class imbalance within each bias group. If this approach was used to debias a classification task on real data, the cause of these performance disparities may have been unclear since it could have been related to other subgroup attributes unaccounted for. Finally, we observed that the saliency scores align with performance of the model for each bias mitigation strategy. With knowledge of the exact regions that should be relevant for model decision-making, and of how a particular explainability method manifests in the absence of the bias effect, this framework facilitates a highly objective interpretation of explainable AI in medical imaging. Further work could use this setup for studying the extent to which different explainability methods highlight regions associated with predictive targets or bias effects, or developing new methods that are better at identifying salient regions of medical images.

Although the SimBA tool enables the generation of increasingly complex dataset bias scenarios, real-world medical imaging data can be even more complex than the synthetic data we analyze in this controlled setup. This is because real datasets inherently contain numerous confounding and interacting biases, many of which may be imperceptible to humans. For instance, it may be necessary to account for disease prevalence and population shift when evaluating performance disparities in real data, since the underlying distribution of other subgroup-associated factors may confound analysis. Glocker et al[16] account for this by performing test set resampling to balance the representation of subgroups for a more accurate representation of algorithmic bias. Conversely, using the SimBA data generation tool in our framework, these underlying distribution shifts can be mitigated, as we do here with stratified sampling, or systematically simulated and evaluated by generating datasets with shifts in the distributions of subject or disease effects. Thus, this analysis framework provides a powerful technique for investigating algorithmic encoding of bias and efficacy of bias mitigation strategies on a range of dataset bias scenarios, in which the biases could be represented by localized spatial deformations, sampled distribution of global morphology, or intensity-based simulated artifacts, to name a few. Therefore, these models and methods can be studied and benchmarked on fully controlled setups prior to making conclusions on real-world data. The control and flexibility over the dataset composition provided by SimBA, and the ability to rigorously test in silico trials of the same simulated "subjects" with different bias manifestations as we present in this framework facilitates a comprehensive and objective exploration of biases in medical imaging AI, which was not previously possible.

This first feasibility study was limited by restricting the scope to simulation of spatially localized morphological bias effects in structural neuroimaging data and analysis of subsequent impacts on a single CNN model architecture. Furthermore, since the biases modeled in this article are not intended to faithfully represent real-world subpopulations, the results are not likely to be directly generalizable to models developed with real-world data. However, using this structured analysis framework as a baseline, we believe that there are numerous avenues for the research community to pursue, which will assist in developing foundational knowledge of how medical imaging-related bias features impact AI models. For instance, the potential relationship between the magnitude of performance disparities and proximity of disease and bias regions warrants further investigation into whether these results are consistent when using vision transformer-like architectures, where the model learns a more complex representation of global and local interactions than what is possible in a standard CNN. Furthermore, this framework could provide particular utility in the development of new bias mitigation strategies (especially unsupervised ones), since the user can ensure that all the sources of bias in the data are being addressed, which is not the case if such a strategy were to be evaluated on real data. Aside from studying other organs, arbitrary biases, and disease effects, clinical knowledge of well-defined pathologies (eg, lesions) could be integrated to more faithfully model real diseases. Additionally, as users define the disease effects, predictive tasks are not limited to classification, but could also be extended to regression or segmentation scenarios.

## Conclusion

In summary, this work introduced a structured framework for analyzing bias manifestation and mitigation in medical imaging AI. The results show that this framework has the potential to greatly improve understanding of bias in medical imaging deep learning pipelines, which alongside studies investigating the impacts of real-world biased data, will support the development of clinical AI tools that are more robust, responsible, and fair.

## Author contributions

Emma A.M. Stanley (Conceptualization, Methodology, Software, Validation, Formal Analysis, Investigation, Data curation, Writing—original draft, Writing—review and editing, Visualization, Funding acquisition), Raissa Souza (Conceptualization, Methodology, Software, Writing—review and editing), Anthony J. Winder (Formal analysis, Writing—review and editing), Vedant Gulve (Software), Kimberly Amador (Methodology, Writing—review and editing), Matthias Wilms (Conceptualization, Methodology, Software, Writing—review and editing, Supervision), and Nils D. Forkert (Conceptualization, Methodology, Writing—review and editing, Supervision, Funding acquisition).

## Supplementary material

Supplementary material is available at *Journal of the American Medical Informatics Association* online.

## Funding

This work was supported by Alberta Innovates, Natural Sciences and Engineering Research Council of Canada, River Fund at Calgary Foundation, Canada Research Chairs Program, University of Calgary Department of Pediatrics, and Alberta Children's Hospital Foundation.

## Conflicts of interest

None declared.

## Data availability

Datasets, code, and train/validation/test splits are available for download from https://github.com/estanley16/SimBA.

*Medical Imaging, and Ethical and Philosophical Issues in Medical Imaging.* Lecture Notes in Computer Science. Springer Nature Switzerland; 2023:194-204. https://doi.org/10.1007/978-3-031-45249-9_19